\documentclass[a4paper]{article}

\pdfoutput=1
\usepackage{INTERSPEECH2018}
\usepackage{multirow}
\usepackage{subcaption}

\title{Unsupervised and Efficient Vocabulary Expansion for\\ Recurrent Neural Network Language Models in ASR}
\name{Yerbolat Khassanov, Eng Siong Chng}
\address{
 Rolls-Royce@NTU Corporate Lab, Nanyang Technological University, Singapore}
\email{yerbolat002@edu.ntu.sg, aseschng@ntu.edu.sg}

\begin{document}

\maketitle
\begin{abstract}
In automatic speech recognition (ASR) systems, recurrent neural network language models (RNNLM) are used to rescore a word lattice or N-best hypotheses list.
Due to the expensive training, the RNNLM's vocabulary set accommodates only small shortlist of most frequent words.
This leads to suboptimal performance if an input speech contains many out-of-shortlist (OOS) words.

An effective solution is to increase the shortlist size and retrain the entire network which is highly inefficient.
Therefore, we propose an efficient method to expand the shortlist set of a pretrained RNNLM without incurring expensive retraining and using additional training data.
Our method exploits the structure of RNNLM which can be decoupled into three parts: input projection layer, middle layers, and output projection layer.
Specifically, our method expands the word embedding matrices in projection layers and keeps the middle layers unchanged.
In this approach, the functionality of the pretrained RNNLM will be correctly maintained as long as OOS words are properly modeled in two embedding spaces.
We propose to model the OOS words by borrowing linguistic knowledge from appropriate in-shortlist words.
Additionally, we propose to generate the list of OOS words to expand vocabulary in unsupervised manner by automatically extracting them from ASR output.

\end{abstract}

\section{Introduction}

\begin{figure*}[ht]
\vspace{-10pt}
\centering
\includegraphics[width=0.9\linewidth]{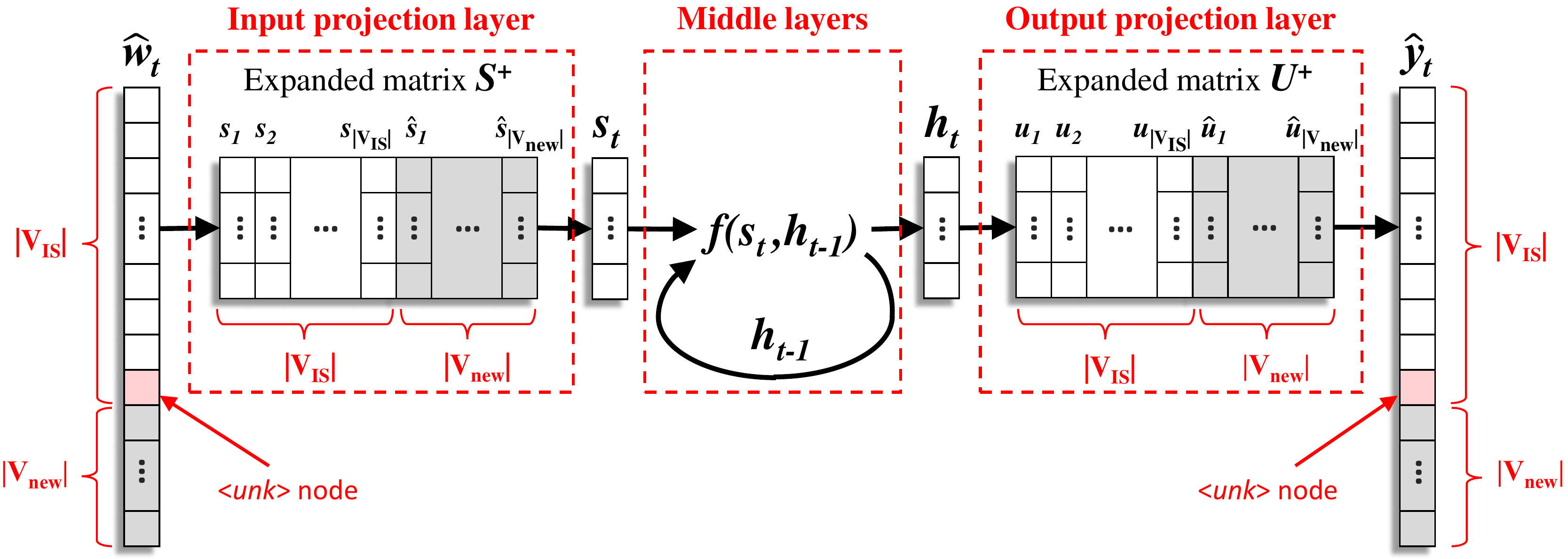}
\caption{RNNLM architecture after vocabulary expansion.}
\label{fig:LSTM}
\vspace{-10pt}
\end{figure*}
The language model (LM) plays an important role in automatic speech recognition (ASR) system.
It ensures that recognized output hypotheses obey the linguistic regularities of the target language.
The LMs are employed at two different stages of the state-of-the-art ASR pipeline: decoding and rescoring.
At the decoding stage, a simple model such as count-based $n$-gram~\cite{goodman2001bit} is used as a background LM to produce initial word lattice.
At the rescoring stage, this word lattice or $N$-best hypotheses list extracted from it is rescored by a more complex model such as recurrent neural network language model (RNNLM)~\cite{mikolov2010recurrent,sundermeyer2012lstm,liu2016two}.

Due to the simplicity and efficient implementation, the count-based $n$-gram LM is trained over a large vocabulary set, typically in the order of hundreds of thousands words.
On the other hand, computationally expensive RNNLM is usually trained with a small subset of most frequent words known as in-shortlist (IS) set, typically in the order of tens of thousands words, whereas the remaining words are deemed out-of-shortlist (OOS) and jointly modeled by single node $<$\textit{unk}$>$~\cite{park2010improved}.
Since the probability mass of $<$\textit{unk}$>$ node is shared by many words, it will poorly represent properties of the individual words leading to unreliable probability estimates of OOS words.
Moreover, these estimates tend to be very small which makes RNNLM biased in favor of hypotheses mostly comprised of IS words.
Consequently, if an input speech with many OOS words is supplied to ASR system, the performance of RNNLM will be suboptimal.

An effective solution is to increase the IS set size and retrain the entire network.
However, this approach is highly inefficient as training RNNLM might take from several days up to several weeks depending on the scale of application~\cite{jozefowicz2016exploring}.
Moreover, additional textual data containing training instances of OOS words would be required, which is difficult to find for rare domain-specific words.
Therefore, the effective and efficient methods to expand RNNLM's vocabulary coverage is of great interest~\cite{chen2015strategies}.

In this work, we propose an efficient method to expand the vocabulary of pretrained RNNLM without incurring expensive retraining and using additional training data.
To achieve this, we exploit the structure of RNNLM which can be decoupled into three parts: 1) input projection layer, 2) middle layers and 3) output projection layer as shown in figure~\ref{fig:LSTM}.
The input and output projection layers are defined by input and output word embedding matrices that perform linear word transformations from high to low and low to high dimensions, respectively.
The middle layers are a non-linear function used to generate high-level feature representation of contextual information.
Our method expands the vocabulary coverage of RNNLM by inserting new words into input and output word embedding matrices, and keeping the parameters of middle layers unchanged.
This method keeps the functionality of pretrained RNNLM intact as long as new words are properly modeled in input and output word embedding spaces.
We propose to model the new words by borrowing linguistic knowledge from other ``similar" words present in word embedding matrices.

Furthermore, the list of OOS words to expand vocabulary can be generated either in supervised or unsupervised manners. 
For example, in supervised manner, they can be manually collected by the human expert.
Whereas in unsupervised manner, a subset of most frequent words from OOS set can be selected.
In this work, we propose to generate the list of OOS words in unsupervised manner by automatically extracting them from ASR output.
The motivation is that the background LM usually covers much larger vocabulary, and hence, during the decoding stage it will produce a word lattice which will contain the most relevant OOS words that might be present in test data.

We evaluate our method by rescoring $N$-best list output from the state-of-the-art TED\footnote{https://www.ted.com/} talks ASR system.
The experimental results show that vocabulary expanded RNNLM achieves $4\%$ relative word error rate (WER) improvement over the conventional RNNLM.
Moreover, $7\%$ relative WER improvement is achieved over the strong Kneser-Ney smoothed $5$-gram model used to rescore the word lattice.
Importantly, all these improvements are achieved without using additional training data and by incurring very little computational cost.

The rest of the paper is organized as follows.
The related works on vocabulary coverage expansion of RNNLMs are reviewed in section 2.
Section 3 briefly describes the RNNLM architecture. 
Section 4 presents the proposed methodology to increase the IS set of RNNLM by expanding the
word embedding matrices. 
In section 5, the experiment setup and obtained results are discussed.
Lastly, section 6 concludes the paper.

\section{Related works}
This section briefly describes popular approaches to expand the vocabulary coverages of RNNLMs.
These approaches mostly focus on intelligently redistributing the probability mass of $<$\textit{unk}$>$ node among OOS words, optimizing the training speed for large-vocabulary models or training sub-word level RNNLM.
These approaches can also be used in combination.

\textbf{Redistributing probability mass of $<$\textit{unk}$>$}:
Park et al.~\cite{park2010improved} proposed to expand the vocabulary coverage by gathering all OOS words under special node $<$\textit{unk}$>$ and explicitly modeling it together with IS words, see figure~\ref{fig:LSTM}.
This is a standard scheme commonly employed in the state-of-the-art RNNLMs.

The probability mass of $<$\textit{unk}$>$ node is then can be redistributed among OOS words by using statistics of simpler LMs such as count-based $n$-gram model as follows:
\vspace{-2pt}
\begin{gather}
\scriptsize
\tilde{P}_{\textit{R}}(w_{t+1}|h_t) = 
\begin{cases}
    P_{\textit{R}}(w_{t+1}|h_t) & w_{t+1}\in\textit{V}_{\textit{IS}}\\
    \beta(w_{t+1}|h_t)P_{\textit{R}}(\mathopen<\textit{unk}\mathclose>|h_t)	& \textit{otherwise}
\end{cases}\\
\beta(w_{t+1}|h_t) =\frac{P_{\textit{N}}(w_{t+1}|h_t)}{\sum_{w\notin \textit{V}_{\textit{IS}}}P_{\textit{N}}(w|h_t)} 
\end{gather}
where $P_{R}()$ and $P_{N}()$ are conditional probability estimates according to RNNLM and $n$-gram LM respectively, for some word $w_{t+1}$ given context $h_{t}$.
The $n$-gram model is trained with whole vocabulary set $V$, whereas RNNLM is trained
with smaller in-shortlist subset $V_{\textit{IS}} \subset V$.
The $\beta()$ is a normalization coefficient used to ensure the sum-to-one constraint of obtained probability function $\tilde{P}_{R}()$.

Later,~\cite{sundermeyer2012lstm} proposed to uniformly redistribute the probability mass of $<$\textit{unk}$>$ token among OOS words as follows:
\vspace{-3pt}
\begin{eqnarray}
\label{eq:sun}
\small
\tilde{P}_{\textit{R}}(w_{t+1}|h_t) =
\begin{cases}
    P_{\textit{R}}(w_{t+1}|h_t) & w_{t+1}\in\textit{V}_{\textit{IS}}\\
    \dfrac{P_{\textit{R}}(\mathopen<\textit{unk}\mathclose>|h_t)}{|V\backslash V_{\textit{IS}}|+1}	& \textit{otherwise}
\end{cases}
\end{eqnarray}
where `$\backslash$' symbol is the set difference operation.
In this way, the vocabulary coverage of RNNLM is expanded to the full vocabulary size $|V|$ without relying on the statistics of simpler LMs. 

\textbf{Training speed optimization:}
Rather than expanding vocabulary of the pretrained model, this group of studies focuses on speeding-up the training of large-vocabulary RNNLMs.

One of the most effective ways to speed up the training of RNNLMs is to approximate the softmax function.
The softmax function is used to normalize obtained word scores to form a probability distribution, hence, it requires scores $y_{w'}$ of every word in the vocabulary:
\begin{eqnarray}
\label{eq:softmax}
softmax(y_w)=\frac{exp(y_w)}{\sum_{w'\in V_{\textit{IS}}}exp(y_{w'})}
\end{eqnarray}
Consequently, its computational cost is proportional to the number of words in the vocabulary and it dominates the training of the whole model which is the network's main bottleneck~\cite{mnih2012fast}.

Many techniques have been proposed to approximate the softmax computation.
The most popular ones include hierarchical softmax~\cite{morin2005hierarchical,mnih2009scalable,mikolov2011extensions}, importance sampling~\cite{bengio2003quick,bengio2008adaptive} and noise contrastive estimation~\cite{mnih2012fast,chen2015recurrent}.
The comparative study of these techniques can be found in~\cite{chen2015strategies,grave2016efficient}.
Other techniques, besides softmax function approximation, to speed up the training of large-vocabulary models can be found in~\cite{mikolov2011strategies}.

\textbf{Sub-word level RNNLM:}
Another effective method to expand the vocabulary coverage is to train a sub-word level RNNLM.
Different from standard word-level RNNLMs, they model finer linguistic units such as characters~\cite{sutskever2011generating} or syllables~\cite{mikolov2012subword}, hence, a larger range of words will be covered.
Furthermore, character-level RNNLM doesn't suffer from the OOS problem, though, it performs worse than word-level models\footnote{At least for English.}~\cite{mikolov2012subword}.
Recently, there has been a lot of research effort aiming to train the hybrid of word and sub-word level models where promising results are obtained~\cite{kim2016character,jozefowicz2016exploring,xu2018neural}.

\section{RNNLM architecture}
The conventional RNNLM architecture can be decoupled into three parts: 1) input projection layer, 2) middle layers and 3) output projection layer, as shown in figure~\ref{fig:LSTM}.
The input projection layer is defined by input word embedding matrix $S\in\mathbb{R}^{d_s\times|V_{\textit{IS}}|}$ used to transform the one-hot encoding representation of  word $w_t\in\mathbb{R}^{|V_{\textit{IS}}|}$ at time $t$ into lower dimensional continuous space vector $s_t\in\mathbb{R}^{d_s}$, where $d_s$ is input word embedding vector dimension:
\vspace{-5pt}
\begin{eqnarray}
s_t=Sw_t
\end{eqnarray}
This vector $s_t$ and compressed context vector from previous time step $h_{t-1}\in\mathbb{R}^{d_h}$ are then merged by non-linear middle layer, which can be represented as function $f()$, to produce a new compressed context vector $h_t\in\mathbb{R}^{d_h}$, where $d_h$ is context vector dimension:
\vspace{-6pt}
\begin{eqnarray}
h_t=f(s_t, h_{t-1})
\end{eqnarray}
The function $f()$ can be simple activation units such as sigmoid and hyperbolic tangent, or more complex units such as LSTM~\cite{sundermeyer2012lstm} and GRU~\cite{chung2015gated}.
The middle layer can also be formed by stacking several such functions.

The compressed context vector $h_t$ is then supplied to output projection layer where it is transformed into higher dimension vector $y_t\in\mathbb{R}^{|V_{\textit{IS}}|}$ by output word embedding matrix $U\in\mathbb{R}^{d_h\times|V_{\textit{IS}}|}$:
\vspace{-6pt}
\begin{eqnarray}
y_t=U^Th_t
\end{eqnarray}
The entries of output vector $y_t$ represent the scores of words to follow the context $h_t$.
These scores are then normalized by softmax function to form probability distribution (eq.~\eqref{eq:softmax}).

\section{Vocabulary expansion}
This section describes our proposed method to expand the vocabulary coverage of pretrained RNNLM.
Our method is based on the observation that input and output projection layers learn the word embedding matrices, and middle layers learn the mapping from the input word embedding vectors to compressed context vectors.
Thus, by modifying the input and output word embedding matrices to accommodate new words, we can expand the vocabulary coverage of RNNLM.
Meanwhile, the parameters of middle layers are kept unchanged which allows us to avoid expensive retraining.
This approach will preserve the linguistic regularities encapsulated within pretrained RNNLM as long as the new words are properly modeled in input and output embedding spaces.
To model the new words, we will use word embedding vectors of ``similar" words present in $V_{\textit{IS}}$ set.

The proposed method has three main challenges:
1) how to find relevant OOS words for vocabulary expansion,
2) criteria to select ``similar" candidate words to model a target OOS word and
3) how to expand the word embedding matrices.
The details are discussed in section~\ref{relevant},~\ref{candidates} and~\ref{matrices}, respectively.

\subsection{Finding relevant OOS words}
\label{relevant}
The first step to vocabulary expansion is finding relevant OOS words.
This step is important as expanding vocabulary with irrelevant words absent in the input test data is ineffective.
The relevant OOS words can be found either in supervised or unsupervised manners.
For example, in supervised manner, they can be manually collected by human expert.
In unsupervised manner, the subset of most frequent OOS words can be selected.

In this work, we employed an unsupervised method where relevant OOS words are automatically extracted from the ASR output.
The reason is that at the decoding stage a background LM covering very large vocabulary set is commonly employed.
Subsequently, the generated word lattice will contain the most relevant OOS words that might be present in the input test data.

\subsection{Selecting candidate words}
\label{candidates}
Given a list of relevant OOS words, let's call it set $V_{\textit{new}}$, the next step is to select candidate words that will be used to model each of them.
The selected candidates must be present in $V_{\textit{IS}}$ set and should be similar to the target OOS word in both semantic meaning and syntactic behavior.
Selecting inadequate candidates might deteriorate the linguistic regularities incorporated within pretrained RNNLM, thus, they should be carefully inspected.
In natural language processing, many effective techniques exist that can find appropriate candidate words satisfying conditions mentioned above~\cite{miller1998wordnet,mikolov2013efficient,pennington2014glove}.

\subsection{Expanding word embedding matrices}
\label{matrices}
This section describes our proposed approach to expand the word embedding matrices $S$ and $U$.

\textbf{The matrix \textit{S}:}
This matrix holds input word embedding vectors $S=[s_1\ \ ...\ \ s_{|V_{\textit{IS}}|}]$ and it's used to transform words from discrete form into lower dimensional continuous space.
In this space, vectors of ``similar" words are clustered together~\cite{collobert2008unified}.
Moreover, these vectors have been shown to capture meaningful semantic and syntactic features of language~\cite{mikolov2013linguistic}.
Subsequently, if two words have similar semantic and syntactic roles, their embedding vectors are expected to belong to the same cluster.
As such, words in a cluster can be used to approximate a new word that belongs to the same cluster.

For example, let's consider a scenario where we want to add a new word Astana, the capital of Kazakhstan, to the vocabulary set of an existing RNNLM.
Here, we can select candidate words from $V_{\textit{IS}}$ with similar semantic and syntactic roles such as London and Paris.
Specifically, we extract the input word embedding vectors of selected candidates  $C=\{s_{london}, s_{paris}\}$ and combine them to form a new input word embedding vector \mbox{$\hat{s}_{astana}\in\mathbb{R}^{d_s}$} as follows:
\vspace{-1pt}
\begin{equation}
\label{eq:new_emb_vector}
\hat{s}_{astana}=\frac{\sum_{s\in C}m_{s}s}{|C|}\\
\end{equation}
where $m_{s}$ is some normalized metric used to weigh the candidates.
We repeat this procedure for all words in set $V_{\textit{new}}$.

Obtained new input word embedding vectors are then used to form matrix $\hat{S}=[\hat{s}_1\ ...\ \hat{s}_{|V_{\textit{new}}|}]$ where \mbox{$\hat{S}\in\mathbb{R}^{d_s\times|V_{\textit{new}}|}$}, which is appended to the initial matrix
\mbox{$S\in\mathbb{R}^{d_s\times|V_{\textit{IS}}|}$} to form the expanded matrix:
\begin{equation}
S^+=[S\ \ \hat{S}]
\end{equation}
where $S^+\in\mathbb{R}^{d_s\times(|V_{\textit{IS}}|+|V_{\textit{new}}|)}$.

The input word vector $w_{t}\in\mathbb{R}^{|V_{\textit{IS}}|}$ should be also expanded to accommodate the new words from $V_{\textit{new}}$, which results in the new input vector $\hat{w}_{t}\in\mathbb{R}^{|V_{\textit{IS}}|+|V_{\textit{new}}|}$.
The input vector and input word embedding matrix after expansion are depicted in figure~\ref{fig:LSTM}.

\textbf{The matrix \textit{U}:}
This matrix holds the output word embedding vectors $U=[u_1\ \ ...\ \ u_{|V_{\textit{IS}}|}]$ where $u\in\mathbb{R}^{d_h}$.
These vectors are compared against context vector $h_t$ using the dot product to determine the score of the next possible word $w_{t+1}$\cite{mnih2012fast}.
Intuitively, for a given context $h_t$, the interchangeable words with similar semantic and syntactic roles should have similar scores to follow it.
Therefore, in the output word embedding space, interchangeable words should belong to the same cluster.
Subsequently, we can use the same procedure and candidates which were used to expand matrix $S$ to model the new words in output word embedding space.
However, this time we operate in the column space of matrix $U$.
The output vector and output word embedding matrix after expansion are depicted in figure~\ref{fig:LSTM}.

\section{Experiment}
This section describes experiments conducted to evaluate the efficacy of proposed vocabulary expansion method for pretrained RNNLMs on ASR task.
The ASR system is built by Kaldi~\cite{povey2011kaldi} speech recognition toolkit on the first release of TED-LIUM~\cite{rousseau2012ted} speech corpus.
To highlight the importance of vocabulary expansion, we train the LMs on generic-domain text corpus One Billion Word Benchmark (OBWB)~\cite{chelba2013one}.

As the baseline LMs, we trained two state-of-the-art models, namely the modified Kneser-Ney smoothed 5-gram (KN5)~\cite{chen1999empirical} and recurrent LSTM network (LSTM)~\cite{sundermeyer2012lstm}.
We call our system VE-LSTM which is constructed by expanding the vocabulary of the baseline LSTM.
The performance of these three models is evaluated using both perplexity and WER.

\textbf{Experiment setup:}
The TED-LIUM corpus is comprised of monologue talks given by experts on specific topics, its characteristics are given in table~\ref{tab:ted}.
Its train set was used to build the acoustic model with the `nnet3+chain' setup of Kaldi including the latest developments.
Its dev set was used to tune hyper-parameters such as the number of candidates to use to model the new words, word insertion penalty and LM scale.
The test set was used to compare the performance of proposed VE-LSTM and two baseline models.
Additionally, the TED-LIUM corpus has a predesigned pronunciation lexicon of $150k$ words which was also used as a vocabulary set for baseline LMs.

The OBWB corpus consists of text collected from various domains including the news and parliamentary speeches.
Its train set contains around $700M$ words and is used to train both baseline LMs.
Its validation set of size $141k$ words was used to stop the training of LSTM model.
\begin{table}[t]
\begin{center}
\caption{\label{tab:ted}The characteristics of TED-LIUM corpus}
\vspace{-5pt}
\begin{tabular}{ l | c | c | c | c }
\hline
\multicolumn{2}{c|}{Characteristics}			& Train		& Dev  	& Test \\ \hline \hline
\multicolumn{2}{l|}{No. of talks}  				& 774		& 8		& 11 \\
\multicolumn{2}{l|}{No. of words}  				& 1.3M		& 17.7k	& 27.5k \\
\multicolumn{2}{l|}{Total duration} 			& 122hr & 1.5hr & 2.5hr \\
\hline
\end{tabular}
\end{center}
\vspace{-20pt}
\end{table}

The baseline KN5 was trained using SRILM~\cite{stolcke2002srilm} toolkit with $150k$ vocabulary.
It was used to rescore the word lattice and 300-best list.
Its pruned\footnote{The pruning coefficient is $10^{-7}$.} version KN5\_pruned was used as a background LM during the decoding stage.

The baseline LSTM was trained as a four-layer network similar to~\cite{sundermeyer2012lstm} using our own implementation in PyTorch~\cite{paszke2017automatic}.
The LSTM explicitly models only the $10k$ most frequent words\footnote{Plus the beginning $<$s$>$ and end of sentence $<$/s$>$ symbols.} of $150k$ vocabulary set.
The remaining $140k$ words are modeled by uniformly distributing the probability mass of $<$\textit{unk}$>$ node using equation~(\ref{eq:sun}).
Thus, the input $w_t$ and output $y_t$ vector sizes are $10k+1$ which we call as $V_{\textit{IS}}$ set.
Hence, the baseline LSTM theoretically models the same $150k$ vocabulary set as KN5.
The OOS rate with respect to dev and test sets are $6.8\%$ and $5.5\%$, respectively.
The input $d_s$ and output $d_h$ word embedding vector dimensions were set to $300$ and $1500$, respectively.
The parameters of the model are learned by truncated backpropagation through time algorithm (BPTT)~\cite{williams1990efficient} unrolled for 10 steps.
For regularization, we applied 50\% dropout on the non-recurrent connections as suggested by~\cite{zaremba2014recurrent}.

The VE-LSTM model is obtained by expanding the vocabulary of baseline LSTM with OOS words extracted from the ASR output.
For example, to construct the VE-LSTM model for the test set, we collect the list of OOS words $V_{\textit{new}}$ from the recognized hypotheses of the test set.
For each OOS word in $V_{\textit{new}}$, we then select the appropriate set of candidate words $C$.
The selection criteria will be explained later.
Next, selected candidates are used to model the new input and output word embedding vectors of target OOS words as in equation~(\ref{eq:new_emb_vector}).
For simplicity, we didn't weigh the selected candidates.
Lastly, these generated new vectors are appended to the input and output word embedding matrices of baseline LSTM model, see figure~\ref{fig:LSTM}.
Consequently, the obtained VE-LSTM will explicitly model $10k+1+|V_{\textit{new}}|$ words, whereas the remaining $140k-|V_{\textit{new}}|$ words are modeled by uniformly distributing the probability mass of $<$\textit{unk}$>$ node using equation~(\ref{eq:sun}).

To select candidate words we used the classical skip-gram model~\cite{mikolov2013efficient}.
The skip-gram model is trained with default parameters on OBWB corpus covering all unique words.
Typically, when presented with a target OOS word, the skip-gram model returns a list of ``similar" words.
From this list, we only select top eight\footnote{This number was tuned on the dev set.} words which are present in $V_{\textit{IS}}$ set.

\textbf{Results:}
The experiment results are given in table~\ref{tab:results}.
We evaluated LMs on perplexity and WER measure.
The perplexity was computed on the reference data and the OOS words for vocabulary expansion were extracted from the reference data as well.
The perplexity results computed on the test set show that \mbox{VE-LSTM} significantly outperforms KN5 and LSTM models by $31\%$ and $18\%$ relative, respectively.

\begin{table}[t]
\begin{center}
\caption{\label{tab:results} The perplexity and WER results on dev and test sets of TED-LIUM corpus}
\vspace{-5pt}
\begin{tabular}{ l | c | c | c | c | c }
\hline
\multirow{2}{*}{LM}		& \multicolumn{2}{|c|}{Perplexity}	& \multicolumn{3}{|c}{WER (\%)}\\\cline{2-6}
						& Dev  	& Test 						& Rescore 	& Dev 		& Test\\\hline\hline
KN5\_pruned				& 384 	& 341 						& -			& 13.05		& 12.82 \\\hline
\multirow{2}{*}{KN5}	& \multirow{2}{*}{237}	& \multirow{2}{*}{218}	& Lattice 	& 10.85 & 10.49 \\
						& 						& 			& 300-best	& 11.03 	& 10.79	\\\hline
LSTM					& 211 	& 185 						& 300-best	& 10.63 	& 10.18	\\\hline
VE-LSTM  				& \textbf{163}	& \textbf{150}		& 300-best	& \textbf{10.44}&\textbf{9.77}  \\\hline
\end{tabular}
\end{center}
\vspace{-20pt}
\end{table}

For WER experiment, the KN5 is evaluated on both lattice and 300-best rescoring tasks. 
The LSTM and \mbox{VE-LSTM} are evaluated only on 300-best rescoring task.
We tried to extract the OOS words for vocabulary expansion from different $N$-best lists.
Interestingly, the best result is achieved when they are extracted from the $1$-best.
The reason is that $1$-best hypothesis list contains high confidence score words, hence, OOS words extracted from it will be reliable.
Whereas using other $N$-best lists will result in unreliable OOS words which confuse the \mbox{VE-LSTM} model.
The VE-LSTM outperforms the baseline LSTM model by $4\%$ relative WER.
Compared to KN5 used to rescore the word lattice, $7\%$ relative WER improvement is achieved.
Such improvements suggest that the proposed vocabulary expansion method is effective.

\section{Conclusion}
In this paper, we have proposed an efficient vocabulary expansion method for pretrained RNNLM.
Our method which modifies the input and output projection layers, while keeping the parameters of middle layers unchanged was shown to be feasible.
It was found that extracting OOS words for vocabulary expansion from the ASR output is effective when high confidence words are selected. 
Our method achieved significant perplexity and WER improvements on the state-of-the-art ASR system over two strong baseline LMs.
Importantly, the expensive retraining was avoided and no additional training data was used.
We believe that our approach of manipulating input and output projection layers is general enough to be applied to other neural network models with similar architectures.

\section{Acknowledgements}

This work was conducted within the Rolls-Royce@NTU Corporate Lab with support from the National Research Foundation (NRF) Singapore under the Corp Lab@University Scheme.

\bibliographystyle{IEEEtran}

\bibliography{mybib}


\end{document}